# The Partial Response Network: a neural network nomogram

Paulo J. G. Lisboa [a], Sandra Ortega-Martorell [a], Sadie Cashman [b], Ivan Olier [a]

[a] *Department of Applied Mathematics, Liverpool John Moores University, Liverpool, UK*
[b] *Department of Built Environment, Liverpool John Moores University, Liverpool, UK*

\* Corresponding author

Email addresses: P.J.Lisboa@ljmu.ac.uk (P Lisboa), S.OrtegaMartorell@ljmu.ac.uk (S. Ortega-Martorell), S.Cashman@2012.ljmu.ac.uk (S. Cashman), I.A.OlierCaparroso@ljmu.ac.uk (I. Olier)

**ABSTRACT**

Among interpretable machine learning methods, the class of Generalised Additive Neural Networks (GANNs) is referred to as Self-Explaining Neural Networks (SENN) because of the linear dependence on explitic functions of the input variables. In binary classification this shows the precise weight that each input contributes towards the logit. The nomogram is a graphical representation of these weights.

We show that functions of individual and pairs of variables can be derived from a functional Analysis of Variance (ANOVA) representation, enabling an efficient feature selection to be carried by application of the logistic Lasso. This process infers the structure of GANNs which otherwise needs to be predefined. As this method is particularly suited for tabular data, it starts by fitting a generic flexible model, in this case a Multi-layer Perceptron (MLP) to which the ANOVA decomposition is applied. This has the further advantage that the resulting GANN can be replicated as a SENN, which enables further refinement of the univariate and bivariate component functions to take place. The component functions are partial responses hence the SENN is a partial response network.

The Partial Response Network (PRN) is equally as transparent as a traditional logistic regression model, but capable of non-linear classification with comparable or superior performance to the original MLP. In other words, the PRN is a fully interpretable representation of the MLP, at the level of univariate and bivariate effects.

The performance of the PRN is shown to be competitive for benchmark data, against state-of-the-art machine learning methods including Gradient Boosting Machines, Support Vector Machines and Random Forests. It is also compared wtih spline-based Sparse Additive Models (SAM) showing that a semi-parametric representation of the GAM as a neural network can be equally as effective as the SAM though less constrained by the need to set spline nodes.

*Keywords*: Interpretability, Generalised Additive Neural Networks, Self-Explaining Neural Networks, Sparse Additive Model, Machine explanation, Multi-Layer Perceptron

# 1 INTRODUCTION

Artificial intelligence has radically increased the accuracy of inferences made from complex data. However, these algorithms are often difficult to understand by users from other domains and can have unknown failure modes. Moreover, models driven by observational data can be difficult to correct for bias and other artifactual effects that may be present in the data. This has generated interest in interpretable models, which may be considered best practice in real-world applications [1].

In many applications the natural amounts of noise present in data constraint the inference of accurate statistical effects, typically restricting them to simpler models with univariate and bivariate components [2]. This can be the case especially for tabular data, the type used in high-stakes decision support for instance in medicine.

Existing approaches to explaining black boxes [1] generally rely on deriving approximations to the models' response function. They include decision trees for either shallow [3] or deep [4] neural networks, and additive feature attribution methods. Recently, three properties were identified which uniquely specify a unique class of explanation models that complies with the generic framework of additive feature attributions using binary feature selection variables [5]. These properties are local accuracy, lack of impact for features missing in the original input, and consistency in the sense that "if a model changes so that some simplified input's contribution increases or stays the same regardless of the other inputs, that input's attribution should not decrease" [5].

This broad class of models has been unified in a single framework that includes Local Interpretable Model-Agnostic Explanations (LIME) [6]. However local approximations for ascertaining feature attributions, such as sensitivity analysis and saliency maps, can be misleading [1]. Alternative approaches to machine explanation also use signal reconstruction to maximise the activation of certain nodes in deep neural networks, for instance with Generative Adversarial Networks [7], or take a completely different perspective and use information theory principles for metric learning which serves to map data structure for nearest neighbour classification and case-based reasoning [8].

We aim to build a model that is its own explanation. This is achieved by using the structural form of Self-Explaining Neural Networks (SENN) [9]. These networks generalise additive models by assembling them out interpretable basis concepts, or atoms. The basis concepts typically include aggregates of the inputs, features extracted from the data using expert knowledge, and prototypes.

To achieve this aim, our approach is guided by the three desiderata for robust interpretability and explainabilty outlined in [9]: explicitness/intelligibility: *"Are the explanations immediate and understandable?";* faithfulness: *"Are relevance scores indicative of "true" importance?";* and stability: *"How consistent are the explanations for similar/neighboring examples?".*

We agree with this set and propose to extend it with two more requirements:

- Parsimony: *"Do the explanatory variables comprise a minimal set?"*
- Consistency: *"How robust are the explanations to perturbations in the data?"*

Parsimony has a long history in statistics and is widely regarded as central to ensuring robust generalisation by retaining only the most informative atomic forms as inputs. Consistency is also fundamental if we attribute relevance to functions of the input variables, which will be our atomic components, and wish quantify their influence in a way that can be understood by the end-user.

The ultimate goal is to enable end-users to integrate the proposed machine learning model into their reasoning processes. This requires meeting the desiderata above with a model that makes clear the exact weight of each input variable or pairs of variables towards the model prediction, not just in a local neighbourhood of a test point but globally over the complete range of the data.

*1.1 Related work*

An early class of models that are interpretable by design, that is to say self-explaining, is Generalized Additive Models (GAMs) [11]. They allow for non-linear components in one or more variables, although they are often applied as linear combination of univariate functions back-fitted to the data with an assumed starting model. More recently, a computationally efficient method to estimate GAMs was proposed, the Sparse Additive Model (SAM) [11]. This model extends the functional ANOVA model by adding $l_1$ regularisation and additional constraints on the model parameters, which is necessary for identifiability, to obtain a unique solution through convex optimisation. It parameterises the component functions in terms of the smoothing matrix. The model can be shown to have persistence [12] meaning that it will reliably find close to the best subset of explanatory variables, including when there are more variables than observations. The results in our paper confirm that the Lasso [13] method is efficient to find the optimal sparse predictors, even in high dimensions.

Generalised Additive Neural Networks (GANNs) have a long history. They were originally proposed [14] because they are their own interpretation. They take the form of GAMs and hence form a natural bridge between machine learning and traditional statistical methods. GANNs have experienced a resurgence of interest [15] both with assumed [16] and unknown [17] link functions. However, there is no systematic method to configure the GANN by identifying the minimal set of input features for a given classification task. This means that GANNs are used with pre-selected features, usually restricted to be univariate since the search space for bivariate functions is large. Nevertheless, this neural network configuration can be useful for obtaining rigorous estimates of statistical measures of importance e.g. in clinical applications, such as the odds ratios for specific effects [17].

The above methods are efficient but fully parametric, thus assuming prior knowledge e.g. about model structure, in contrast with neural networks which are generic flexible models.

In order to infer the model structure, we start with a pre-trained machine learning model. In this respect, the approach taken in this paper has parallels with a previous derivation of an interpretable machine learning model using Support Vector Machines [18]. In that paper, the derived nomograms are obtained by applying a Taylor expansion to the Gaussian kernels, which are then re-shaped by separately summing the univatiate and bivariate terms. Feature selection is by iterative application of the standard kernel trick to a re-weighted objective function with $l_1$ regularisation.

*1.2 Novel contribution*

The first novel contribution of this paper is to position the class of partial response models which includes [18] as a subset of functional Analysis of Variance (ANOVA) decompositions [19] anchored at a suitable point which we choose to be the median of the data. The functional decomposition makes explicit the different component functions inherent in the original multivariate predictive model, representing it by a finite sum of functions of fewer variables, which are easier to interpret. This has the important benefit that this exact representation can be trunctated in order to remove high-order interactions that may not be reliably estimated and so are treated as noise.

At a more practical level, this paper proposes a computationally efficient method to compute the nomogram for a high performing SENN. This addresses the unsolved problem of configuring the SENN directly from data, removing the requirement for prior feature selection. In common with [18], partial responses are derived from a pre-trained model. However, we model the logit of a MLP, instead of Gaussian kernels from a Support Vector Machine (SVM). This proves to be more stable and reliably derives sparse models with high classification performance.

Third, the proposed method is empirically validated by benchmarking against state-of-the-art classifiers including Gradient Boosting Machines, Support Vector Machines and Random Forests, as well as a spline-based sparse additive model [11].

The main limitation of the method is that it applies to data provided in a rectangular table, that is to say tabular data as opposed to data such as images, speech and text, all of which required additional modelling of their internal structure.

## 2 METHOD

A natural quantity to infer in binary classification is the logit, which is the inverse of the sigmoid link function. This is because the logit is not bounded above and below, unlike probabilities, hence it does not have to be strongly non-linear. In many useful models the logit is close to linear [2]. This makes it easier to estimate it accurately from noisy data.

We start by estimating the probability density function of the posterior distribution of class membership with an appropriately regularised MLP. In the examples shown later, we choose to use a Bayesian framework known as Automatic Relevance Determination (ARD) as this shrinks the hidden layer weights according to the influence of each variable, thus implementing soft variable selection [20]. This is helpful since feature selection is important for the next steps in the method.

The key point of the paper is that the low-order dependencies in the probability densty function can extracted with the functional ANOVA decomposition [19]:

$$logit(P(C|x)) \equiv \varphi(0) + \sum_i \varphi_i(x_i) \\ + \sum_{i \neq j} \varphi_{ij}(x_i, x_j) + \cdots \\ + \sum_{i_1 \neq \cdots \neq i_d} \varphi_{i_1 \ldots i_d}(x_{i_1}, \ldots, x_{i_d}). \quad (1)$$

This decomposition comprises a finite number of terms up to interactions of dimension $d$, each term having all of the variables fixed except those indexed by $i, j, \ldots i_d$. The different tems are orthogonal in a functional sense [19] so can be regarded as independent inputs to a later model.

It is worth noting that (1) is not an equation, but an identity. Therefore, it applies everywhere in the space of possible inputs. It does not represent an expansion that is valid only in the neighbourbood of the anchor point $0$, but rather it is a decomposition of the original function into $2^d$ components, most of which have a much simpler functional form than the original logit function, such that the exact identy holds everywhere. This makes (1) the basis for a globally interpretable model.

Since the ANOVA representation exactly matches the output of the original MLP, the universal approximation capability of the MLP is, in principle, retained. However, for interpretability we make the hypothesis that it is possible to truncate (1) and retain only univariate and bivariate terms without compromising performance.

This is an apparent contradiction of the usual statement that intepretability necessarily leads to performance deficits compared with black box models. We argue that the form of the ANOVA decomposition pulls out the terms that are easier to estimate accurately and by eliminating noisy high-order terms we can potentially enhance classification performance, in a manner analogous to removing principal components with smaller eigenvalues i.e. by retaining the signal while reducing noise.

Therefore, the ANOVA decomposition provides the basis for a computationally efficient algorithm to derive the structure of the GANN. Using a neural network approach is a practical and flexible alternative to parametric modelling of the component functions, such as Sparse Additive Models (SAMs) [11]. In common with SAM, hard feature selection uses a scalable method, which is the Lasso [13].

Thus, the proposed method, whose implementation is detailed later, involves the application of the Lasso to identify the component functions from the ANOVA decomposition that are statistically significant to fit the logit function to the available data. Having identified the most predictive subset of partial responses derived from the MLP, we utilise the structure of a GANN to replicate the output of the truncated ANOVA decomposition, enabling a second iteration to be applied to re-calibrate the component functions, now free of interactions with uninformative variables since they have been removed from the model. In the results section, the second step is shown empirically to further smooth the non-linear component functions, potentially improving classification performance beyond that of the original MLP.

**2.1 Bayesian framework for soft feature selection with ARD**

Neural networks require regularisation methods to smooth the decision functions and so avoid overfitting. The Bayesian framework replaces early stopping or cross-validation by the use of an objective function to regularise the neural network. This involves an interative process where the MLP is first trained to convergence with small initial values for the regularisation hyperparameters. After each iteration, the hyperparameters are updated and the MLP continues training, until all the parameters in the network reach stable values.

For the MLP a typical regulariser is weight decay, which corresponds to $l_2$ regularisation. In Automatic Relevance Determination the aim is to obtain an analytical expression for the strength of the weight decay parameters, of which there is a separate one for all of the weights linked to each input node, plus one for the bias terms in each layer and one more for the output weights. This mirrors the approach used in the group Lasso [21].

A suitable framework was proposed in [20]. The main principle is to consider the neural network weights $\{w\}$ to be distributed, rather than taking point values.

We start with a given data set to be fitted $D = \{x_m, t_m\}$, $m=1 .. N$, comprising observation vectors $x_m$ and targets consisting in this case of binary class labels $\{t_m\}$. The purpose of the classifier is to estimate the probability that $t=1$. Importantly, the estimate is itself uncertain and the Bayesian framework extends to producing estimates of confidence intervals for the predictions made for each observation, including for hold-out data. For the sake of brevity, this aspect of the model is not pursued further in this paper.

The basis of the method is now outlined, denoting by $H$ the model hypothesis given by structure of the MLP, in particular the number of layers, nodes in each layer and choice of activation functions, and denoting by $\alpha$ the set of weight decay strengths described above.

The practical implementation of the method relies on representing the posterior probability for the weights by making use of Bayes theorem:

$$P(w|D, \propto, H) = \frac{P(D|w, \propto, H)P(w|\propto, H)}{Z_W(\alpha)} \quad (2)$$

where the normalising constant represents the integral of the numerator over the possible values of the weights, hence $Z_W(\alpha) = P(D|\propto, H)$.

There is a clear meaning to the two terms in the numerator. The first term measures the fit to the data by the MLP with the set structure and weight values, hence

$$P(D|w, \propto, H) = e^{-S_w} \quad (3)$$

where

$$S_w = -\sum_{m=1}^{N}(t_m log(y_m) + (1-t_m)log(1-y_m)) \quad (4)$$

is the usual cross-entropy of the model outputs $\{y_m\}$ against the target values, employed in statistical models as in probabilistic machine learning models.

The second term represents the assumption, sometimes referred to as *Occam's Razor,* that the best performing model in generalisation to unseen data that are idependent and identically distributed to the training data, will come from the simplest model to acrruately fit the data. In practice this will be the simplest model with comparable performance to the best that is achievable on test data. This assumption is expressed by a prior distribution of the weights that is controlled by its variance. It is well known that the distributional form with maximum entropy for a given variance is the normal distribution, hence

$$P(w|\alpha, H) = e^{-E(w,\alpha)} \quad (5)$$

where the weights are in *K* groups with $N_k$ connection weights in each:

$$E(w, \alpha) = (1/2) \sum_{k=1}^{K} \alpha_k \sum_{l=1}^{N_k} w_{kl}^2. \quad (6)$$

During training the values of the weight decay hyperparameters $\alpha_k$ for least informative groups of variables will rise steadily, compressing the corresponding weigth values towards zero. This is what is referred to as soft pruning. A further stage of hard pruning will be required, where the form of regularisation will switch from $l_2$ to $l_1$.

The hyperparmeters $\alpha_k$ are estimated by maximising their posterior probability,

$$P(\alpha|D, H) = \frac{P(D|\propto, H)P(\alpha|H)}{P(D|H)}. \quad (7)$$

This time a flat prior is assumed and since the denominator is independent of the $\alpha_k$, it is sufficient to maximise $Z_W$ hence

$$P(D|\propto, H) = \int \frac{e^{-S_w - E(w,\alpha)}}{Z_w(\alpha)} dw \quad (8)$$

where the functional form of the normalising factor is required

$$Z_W(\alpha) = \prod_{k=1}^{K}(2\pi/\alpha_k)^{N_K/2}. \quad (9)$$

This is not analytical but it can be reliably estimated by a Laplace approximation

[19] which involves a Taylor expansion of $S(w, \alpha) = -S_w - E(w, \alpha)$ about the current operating point $S^*$ located at the current point values of the weights, referred to as the most probable weights for the current instance of the network, $w^{MP}$, by evaluating the Hessian $A = \nabla\nabla S(w^{MP}, \alpha)$:

$$S^*(w, \alpha) \approx S(w^{MP}, \alpha) + \frac{1}{2}(w - w^{MP})A(w - w^{MP}). \quad (10)$$

This uses the fact that when the MLP has converged, the cost function is at an extremum hence the first derivative vanishes. Assuming uniform, uninformative priors for the hyperparameters, the posterior probability can now be expressed analytically:

$$P(\alpha|D, H) \propto \frac{exp(-S(w^{MP}, \alpha))}{Z_W(\alpha)} (2\pi)^{N_K/2} det(A)^{-1/2}. \quad (11)$$

Maximising (11) results in closed-form estimates of the hyperparameters:

$$\gamma_k = N_k - \alpha_k Tr_k(A^{-1}) = \frac{N_k \sum_{l=1}^{N_k}(w_{kl}^{MP})^2}{N_k \sum_{l=1}^{N_k}(w_{kl}^{MP})^2 + Tr_k(A^{-1})} \quad (12)$$

$$\frac{1}{\alpha_k} = \frac{\sum_{l=1}^{N_k}(w_{kl}^{MP})^2}{\gamma_k} = \frac{\sum_{l=1}^{N_k}(w_{kl}^{MP})^2 + Tr_k(A^{-1})}{N_k} \quad (13)$$

where $Tr_k(A^{-1})$ is the trace of the inverse Hessian for the $N_k$ weights that share a common $\alpha_k$. The interpretation of $\gamma_k$ is the bumber of well-determined parameters in that set of weights. The distribution of weight values generates a corresponding distribution of neural network outputs. In this paper, we use this framework for regularisation and make inferences based on the most likely values of the weights.

The Bayesian framework for Automatic Relevance Determination was implemented in Netlab [22].

## 2.2 The Partial Response Network

This section details the implementation of the Partial Response Network. All data are normalised to unit standard deviation and have the median shifted to zero.

*i. Fit the data with a binary classifier*

The first step is to fit a probabilistic binary classifier to the data in the usual way. We choose to use an MLP with Automatic Relevance Determination because this is computationally efficient and already implements soft feature extraction, as explained in the previous section.

*ii. Apply the ANOVA decomposition to the output of the binary classifier*

The second step is to calculate the terms in the ANOVA decomposition. The component functions in the ANOVA decomposition (1) are the partial responses. Each term is (1) is computed from the logit of the output of the MLP, namely the *logit(P(C|x))* which is the log-odds of the probability of class membership for a given input vector *x*, as follows:

$$\varphi(0) = logit\big(P(C|0)\big) \tag{14}$$

$$\varphi_i(x_i) = logit\big(P(C|(0,..,x_i,..,0))\big) - \varphi(0) \tag{15}$$

$$\varphi_{ij}(x_i, x_j) = logit\Big(P\big(C|(0,..,x_i,..,x_j,..0)\big)\Big) \\ - \varphi_i(x_i) - \varphi_j(x_j) - \varphi(0) \tag{16}$$

The general form of the terms in (1) is a recursive function of nested subsets of the covariate indices $\{i_1, \ldots, i_n\}$:

$$\varphi_{i_1 \ldots i_n}(x_{i_1}, \ldots, x_{i_n}) = logit\big(P(C|x_{i_1}, \ldots, x_{i_n})\big) \\ - \sum_{\{i_1 \neq \cdots \neq i_{n-1}\}} \varphi_{i_1 \ldots i_{n-1}}(x_{i_1}, \ldots, x_{i_{n-1}}) \tag{17}$$

It is straightforward to show that the terms generated by the recursion (17) add up exactly to the logit function as stated in identity (1).

*iii. Apply the logistic Lasso to select terms from the ANOVA decomposition*

The standard Lasso for logistic regression [21] carries out hard feature selection with $l_1$ regularisation. This algorithm is applied with its inputs corresponding, not to the original data vectors but using instead the partial responses defined by (14)-(16). That is to say, for each row in the data, the partial responses are obtained by setting some or all of the other variables to zero and recalculating the response function of the MLP for the modified input vector, then deriving its logit.

This is much the same procedure that is sometimes used to interpret non-linear multivariate functions by visualising the form of the function by varying only one or two variables at a time, anchored on the median of the data. The novel component here is the realisation that this simple method actually computes terms from a functional decomposition that applies globally over the totality of input space and not just locally in the neighbourhood of the anchor point. In statistical terms, each partial response represents an independent effect on the logistic model.

*iv. Construct a GANN/SENN to replicate the output of the logistic Lasso*

As the partial responses are obtained from the output of a neural network i.e. from the weights of the original MLP, it is possible to replicate the feedforward structure of the logistic Lasso with only the selected partial response functions, by using replicates of the original weights in the form of a GANN/SENN.

The replication of the weights is explained below, starting with the set of weights $\{w_{ij}, v_j\}$ and bias terms $\{v_j, v_0\}$ of an initial MLP with $i$ input nodes and $j$ hidden nodes. This is illustrated in fig. 1(a). Recall that the partial responses have been selected by the Lasso with the surviving terms carrying coefficients that we will call, respectively, $\beta_i$ for the univariate associated with input $X_i$ and $\beta_{ij}$ for bivariate response associated with the pair of inputs $\{X_i, X_j\}$. The intercept of the logistic regression Lasso is $\beta_0$. The resulting structured neural network is shown schematically in fig. 1(b) with a univariate response on $x_1$ and a bivariate response on $(x_2, x_3)$.

Only two types of structure need to be represented, one for univariate responses and the other for bivariate responses.

1) Univariate partial response corresponding to input $X_i$

   This is shown in fig.1. (b) for input $X_1$. Zero inputs for all other inputs will not contriubute to the activation of the hidden nodes. The hidden layer weights $w_{1j}$ connected to node $X_1$ remain the same as in the original MLP but the weights and bias to the output node need to be adjusted as follows:

   $$v_j \rightarrow \beta_i * v_j \qquad (18)$$

   $$v_0 \rightarrow \beta_i * \left(v_0 - logit(P(C|0))\right) \qquad (19)$$

2) Bivariate partial response for input pair $\{X_k, X_l\}$

   This is shown in fig.1. (b) for inputs $X_2$ and $X_3$. This time, in order to replicate the partial response multiplied by the Lasso coefficient, it is necessary to add three elements to the structure, namely, a univariate partial response for each of the inputs involved and a coupled network that both inputs feed into together. We will use the generic input indices $k$ and $l$ to avoid confusion with the hidden node index $j$. The hiddent layer weights once again remain unchanged from the original MLP. The output layer weights and biast for the network structure representing the univariate term associated with input $k$ (and similarly for input $l$) are adjusted by

   $$v_j \rightarrow (\beta_k - \beta_{kl}) * v_j \qquad (20)$$

   $$v_0 \rightarrow (\beta_k - \beta_{kl}) * \left(v_0 - logit(P(C|0))\right) \qquad (21)$$

   whereas the weigths and bias for the coupled network are changed according to:

   $$v_j \rightarrow \beta_{kl} * v_j \qquad (22)$$

   $$v_0 \rightarrow \beta_{kl} * \left(v_0 - logit(P(C|0))\right). \qquad (23)$$

3) Finally, an amount is added to the sum total of the values calculated for the bias term in the structured neural network. This amount is equal to the intercept of the logistic Lasso, $\beta_0$.

Two matters are worthy of note. First, if an individual variable is present in both univariate and bivariate effects then the weights for the univariate networks are added, since the logit is a linear function of the hidden node activations.

Second, when a bivariate effect is identified by the PRN witbout an univariate effect being present, it is still possible in the next gradient descent step for the new weights to decouple the bivariate terms, so creating a univariate effect if indeed this is what provides the best fit to the data once the weights of the GANN are re-calibrated without having to adjust for the effects of potentially noisy variables that were removed by the Lasso.

*v. Continue training the GANN with gradient descent*

This is the final step in the calculation of the PRN model. We typically used scaled gradient descent with weight decay to continue training the weights of the structured network in fig. 1. (b), with back-propagation of the error from the common output node through to the output and hidden layer weights, which will now differ. This will have the effect of making usually small changes to the partial responses, which will be shown in the next section. The important aspect of this stage in the implementation of the model is that it will optimise the performance of the final neural network without the effect of the additional, uninformative variables, and also the influence of higher-order interactions that are inherent in the initial fully connected MLP.

*vi. Application of the logistic Lasso to the PRN*

An optional further stage is to apply the Lasso to the partial responses derived from the final PRN model. This can remove effects that, after re-training, are no longer statistically significant for fitting the data.

It is observed in practice that PRN models are very stable and yield almost identical partial responses when the seed of the initial MLP is randomly changed. Likewise, the structure of the PRN in terms of the selected effects, and even more so after the further application of the Lasso, tends to be very consistent.

This provides an interesting parallel between neural networks and statistical linear-in-the-parameters models, which have convex optimisation surfaces and therefore a single optimum. Despite the MLP having from multiple local minima, reducing its form by selecting low-dimensional component functions can render it very stable. Moreover, the linear form of the PRN also makes it interpretable.

## 3 EXPERIMENTAL RESULTS

This section benchmarks the classification performance of the PRN against that of three state-of the-art machine learning models, namely Gradient Boosting Machines [23], Support Vector Machines [24] and Random Forests [25]. The results are compared also with those from SAM, a flexible sparse additive model that is estimated using the backfitting algorithm [10].

The performance of PRN models is demonstrated by application to five data sets from the UCI repository [26]. The data sets were selected because they are widely studied, including by related work in [11], hence the optimal classification performance is known, and also because the vast majority of published work involving these data sets uses black box models for which the form of the dependence on covariates that are essential to fit the data is not apparent. Since the PRN is a probabilistic model there is no need to balance the data, as the cut-point for binary classification can be changed. In any case, we focus on a ranking measure, AUROC.

A more detailed description of the data sets follows.

I. <u>Pima diabetes</u> [27]: This data set comprises measurements recorded from 768 women at least 21 years old, of Pima Indian heritage, tested for diabetes using World Health Organization criteria. One of the variables, Blood Serum Insulin, has significant amounts of missing data. These rows were removed along with all entries with missing values of Plasma Glucose Concentration in a tolerance test, Diastolic Blood Pressure (BP), Triceps Skin Fold Thickness (TSF) or Body Mass Index (BMI), resulting in a reduced dataset with n=532. In line with common practice a subset was randomly selected for training (n=314) and the remaining used for testing (n=268). The additional variables available are Age, Number of Pregnancies and Diabetes Pedigree Function (DPF), a measure of family history of diabetes. The data was z-scored, and the median was shifted to zero. A binary target variable indicated whether or not the individual is diabetic, with a prevalence of 35.7%.

II. <u>German Credit Card</u> [28]: For this dataset, we used the numerical version produced by Strathclyde University, which contains 1000 instances and 24 attributes. A subset of 700 samples with a prevalence of 29.6% were used for training, leaving the remaining samples for test (n=300).

III. <u>Ionosphere</u> [29]: This radar data was collected by a system in Goose Bay, Labrador. This system consists of a phased array of 16 high-frequency antennas with a total transmitted power on the order of 6.4 kilowatts. The targets were free electrons in the ionosphere. 'Good' radar returns are those showing evidence of some type of structure in the ionosphere. 'Bad' returns are those that do not; their signals pass through the ionosphere. Received signals were processed using an autocorrelation function whose arguments are the time of a pulse and the pulse number (17 for the Goose Bay system). Instances (n=351) are described by 2 attributes per pulse number, hence the number of attributes is 34 (all continuous). The task is a binary classification of 'good' or 'bad' radar returns (according to the previous definition). In this study, we removed attribute 2 as it only contained zeros, the data was re-scaled to range [-1; 1], and we used 200 returns for training, with a prevalence of 50.5%, and the remaining samples for test, as in [28].

IV. <u>Wisconsin Breast Cancer – Original (WBC-Original)</u> [30]: This dataset, gathered during 1989 – 1992, records measurements for breast cancer cases such as clump thickness, uniformity of cell size and cell shape, marginal adhesion, single epithelial cell size, bare nuclei, bland chromatin, normal nucleoli, and mitoses (9 attributes). Instances with missing values (such as in variable bare nuclei) were removed as per the literature [31,32], with the new dataset containing 683 instances. In line with those studies, the first 400 instances in the new dataset were used for training set and the remaining (n=283) for test, with a prevalence of 43%. The task is a binary classification of 'benign' or 'malignant'.

V. <u>Wisconsin Breast Cancer – Diagnostic (WBC-Diagnostic)</u> [33]: This dataset was included in the UCI repository in November 1995. The features are computed from digitized images of a fine needle aspirate (FNA) of a breast mass, which describe characteristics of the cell nuclei present in the images. Ten real-valued features are computed for each cell nucleus: radius (mean of distances from centre to points on the perimeter), texture (standard deviation of grey-scale values), perimeter, area, smoothness (local variation in radius lengths), compactness (perimeter^2 / area - 1.0), concavity (severity of concave portions of the contour), concave points (number of concave portions of the contour), symmetry, fractal dimension ("coastline approximation" - 1). The mean, standard error, and "worst" or largest (mean of the three largest values) of these features were computed for each image, resulting in 30 features. Just over the half of the samples available were used for training (n=285) with a prevalence of 50.9%, leaving the remaining for test (n=284). The data was re-scaled to range [0; 1] and the task is also a binary classification of 'benign' or 'malignant'.

The benchmarking results are summarised in Table 1. In all cases, the accuracy measured by the area under the receiver operating characteristic curve (AUROC) averaged over ten random initialisations is comparable with state-of-the-art classifiers, although the partial response networks (PRNs) use fewer variables and are intuitive to interpret. This is the case for PRN following re-training with gradient descent partial responses and following an additional re-calibration (PRN-Lasso).

TABLE 1
PRN AVERAGE PERFORMANCE

| AVERAGE AUROC % (STANDARD DEVIATION %) | MLP | PRN | PRN-LASSO | GBM | SVM | RF |
|---|---|---|---|---|---|---|
| PIMA | 89.9 (3.6) | 89.9 (0.6) | 90.0 (0.6) | 89.7 (0.5) | 87.3 (0.5) | 88.8 (0.3) |
| GERMAN CREDIT CARD | 81.1 (1.5) | 81.3 (0.4) | 81.2 (0.5) | 80.9 (0.4) | 81.1 (0.1) | 80.6 (0.2) |
| IONOSPHERE | 94.6 (1.4) | 98.1 (0.4) | 98.1 (0.4) | 98.6 (0.4) | 99.7 (0.0) | 99.1 (0.1) |
| WBC-ORIGINAL | 99.8 (0.0) | 99.6 (0.1) | 99.7 (0.1) | 99.8 (0.0) | 99.7 (0.0) | 99.8 (0.0) |
| WBC-Diagnostic | 99.3 (0.0) | 99.3 (0.2) | 99.4 (0.0) | 98.9 (0.2) | 99.3 (0.0) | 99.1 (0.1) |

Average performance of the Partial Response Network compared with Gradient Boosting Machine (GBM), Support Vector Machines (SVM) and Random Forests (RF)

Comparing the outputs of the PRN models and alternative methods in Table 1, tested on the same data and measuring statistical significance with the McNemar test, the performance difference was not significant at the 5% level, for all data sets.

Typical partial responses for the Pima diabetes data are shown in fig. 2. The same four variables are always selected and the partial responses are very stable. A fifth variable was selected in 4/10 initialisations. This was the variable Number of Pregnancies, for which the partial response is monotonic increasing and has a smaller overall contribution to the logit than the other four variables selected.

The German Credit Card data set showed a clear structure of models with the first three attributes always selected and AUROC of 0.789 [0.729, 0.849] for both the PRN and PRN-Lasso. The next stable model added Attributes 6,17 and 19, but its performance increased little to AUROC typically 0.791. The addition of a further seven variables increased the AUROC to 0.813, reported in Table 1. This is still well within the confidence interval of the three-variable model. The two partial responses shown in fig. 3 are typical of variables with multi-category attributes.

Eight out of ten initialisations returned the same model for the Ionosphere data, comprising attributes 1, 3, 5, 8 and 27, whose responses are shown in fig. 4.

The two examples of histological data for breast cancer have the expected monotonic responses, shown in figs. 5-6. For the original WBC data, the variables clump thickness, marginal adhesion and bare nuclei alone account for AUROC 0.992 [0.976,1.000], which increases marginally with the inclusion of uniformity of cell shape to between 0.993-0.996 depending on the initialisation. The reported models have six variables for complete stability, by adding normal nucleoli and mitoses.

Similarly, in the WBC diagnostic data worst texture, worst area and worst concavity account for AUROC 0.991 [0.975,1.00] with two other variables contributing marginally to performance, namely area and smoothness.

The performance of the models illustrated in the above figures is listed in Table 2. In direct comparison with the SAM, the PRN matches or outperforms SAM on all data sets and the PRN often selects fewer variables than SAM in its default implementation (https://cran.rproject.org/web/packages/SAM/SAM.pdf).

TABLE 2
PRN AND SAM PERFORMANCES FOR SELECTED MODELS

| AUROC % | MLP | PRN | PRN-LASSO | SAM (SAME FEATURES AS PRN) | SAM (WITH FEATURE SELECTION) |
|---|---|---|---|---|---|
| PIMA | 90.1 | 90.4 | 90.6 | 86.4 | 85.9 |
| GERMAN CREDIT CARD | 80.6 | 82.2 | 82.4 | 76.6 | 77.1 |
| IONOSPHERE | 94.8 | 98.2 | 98.3 | 96.0 | 98.5 |
| WBC-ORIGINAL | 99.8 | 99.7 | 99.8 | 97.9 | 99.5 |
| WBC-Diagnostic | 99.3 | 99.3 | 99.4 | 98.7 | 98.5 |

The two methods agree on glucose and age as core variables to model the diabetes data. The SAM includes also the number of pregnancies whereas the PRN selects primarily BMI and DPF. In the case of the credit card data, the PRN is sparser and more accurate. For the inonosphere data SAM selects a very large number of variables with similar discrimination performance as the PRN and this proliferation of variables occurs also for both WBC data sets. Overall, the two sets of variables are compatible but arguably the PRN is closer to a minimal feature set.

The scalability and power of the method can be further illustrated using an additional data set, the Statlog Shuttle data from the UCI data repository. This is widely studied usually by classifying on all 9 variables, with the aim to achieve an AUROC of 0.999 [37]. With partial responses this accuracy is achieved with two variables, whose dependence is shown in fig. 7. The sufficiency of the model is clear from fig. 8. It also serves to illustrate the capability of the PRN to fit 2-way interactions, which is now apparent from the earlier data sets.

## 4   Discussion and Conclusion

We propose the application of functional ANOVA decomposition followed by the Lasso as a computationally efficient method to derive the structure of a GANN, starting from a probabilistic binary classifier for which we choose a fully connected MLP.

Furthermore, we re-train the GANN to optimise classification performance and also the estimation of the partial responses, which are key to interpretation. The resulting model is the Partial Response Neural Nework. It is a self-explainined neural network (SENN) because of its additive form.

The proposed method carries out feature selection in two stages: first soft feature selection with MLP-ARD, followed by elimination of partial responses using the group Lasso. This is therefore a different approach to using $l_1$ regularisation for the purpose of configuring an MLP [34-35], as neither method uses the log-likelihood function for classification and, while both aim at feature selection, neither is in the form of a GANN.

The results show that the PRN can overcome the interpretability-performance trade-off compared with black box models. We believe that this is due to efficient removal of noisy variables. In deriving a model that is transparent, sparse and consistent, the PRN achieves the desiderata for interpretability outlined in the introduction.

Further work will include quantifying the uncertainty in the estimates of the partial responses and posterior probability of class membership, and also to extend the proposed approach to other multivariate probabilistic binary classifiers.

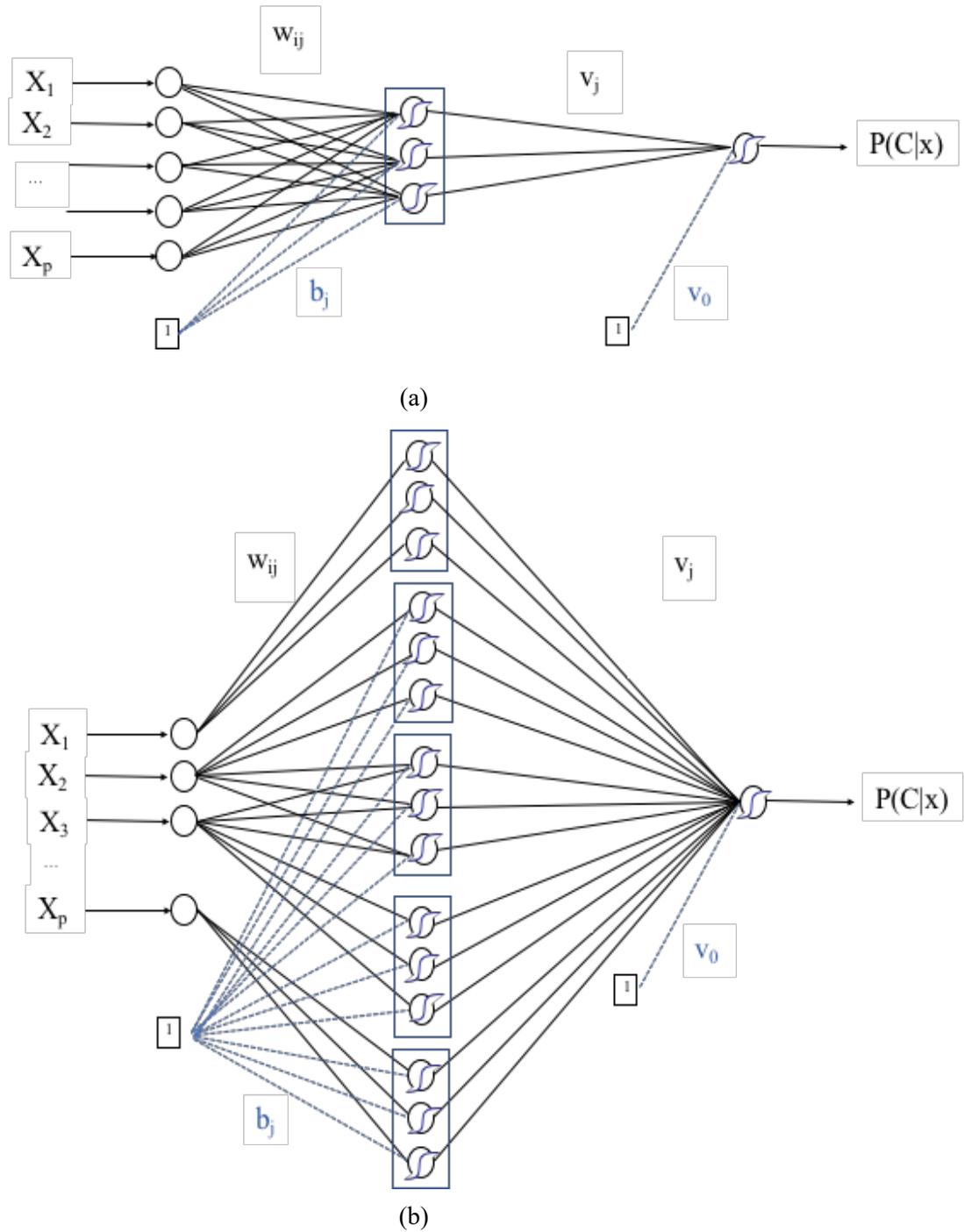

Fig. 1. Structure of (a) the initial MLP and (b) the derived Partial Response Network (PRN). The PRN has the structure of a Generalised Additive Neural Network. It comprises modular replicates of the relevant weights from the original MLP for each univariate or bivariate response retained by the Lasso, adjusted by eqs. (18-23) to initialise the PRN with the exact functional response of the Lasso.

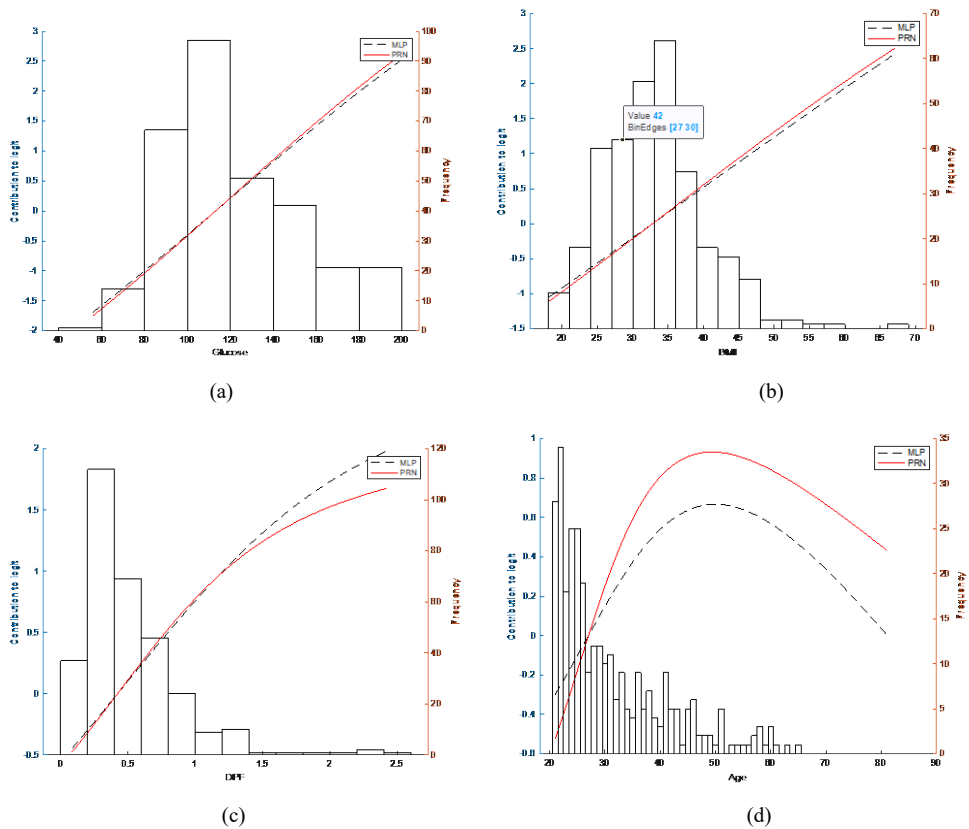

Fig. 2. Partial responses for the Pima diabetes data, overlapped with the histogram of the corresponding variables (right-hand axis). Shown are the responses measured after the second application of gradient descent (solid lines) and from the original MLP (dashed lines). The variables are selected in descending order of maximum contribution to the logit.

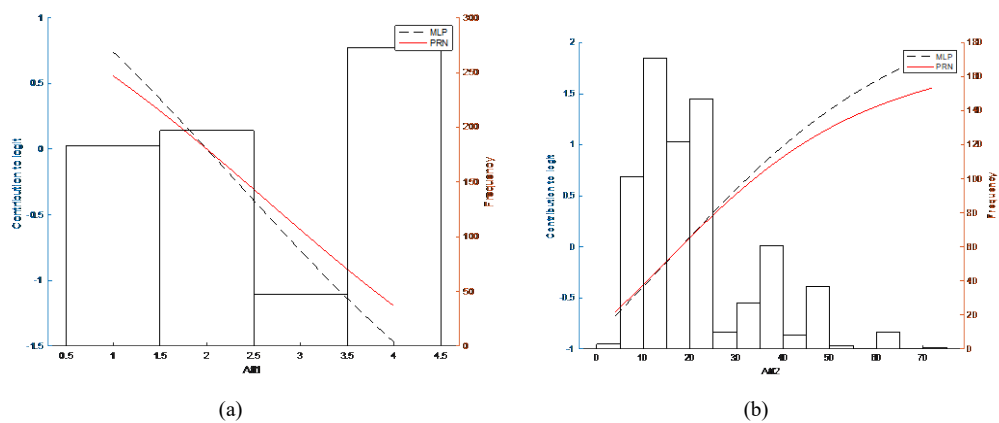

Fig. 3. Partial responses for the German Credit Card data set. Only responses with min-max range >1 are shown. Note that the plots are monotonic either reducing or increasing risk with rising values of the first and second attribute categories, respectively.

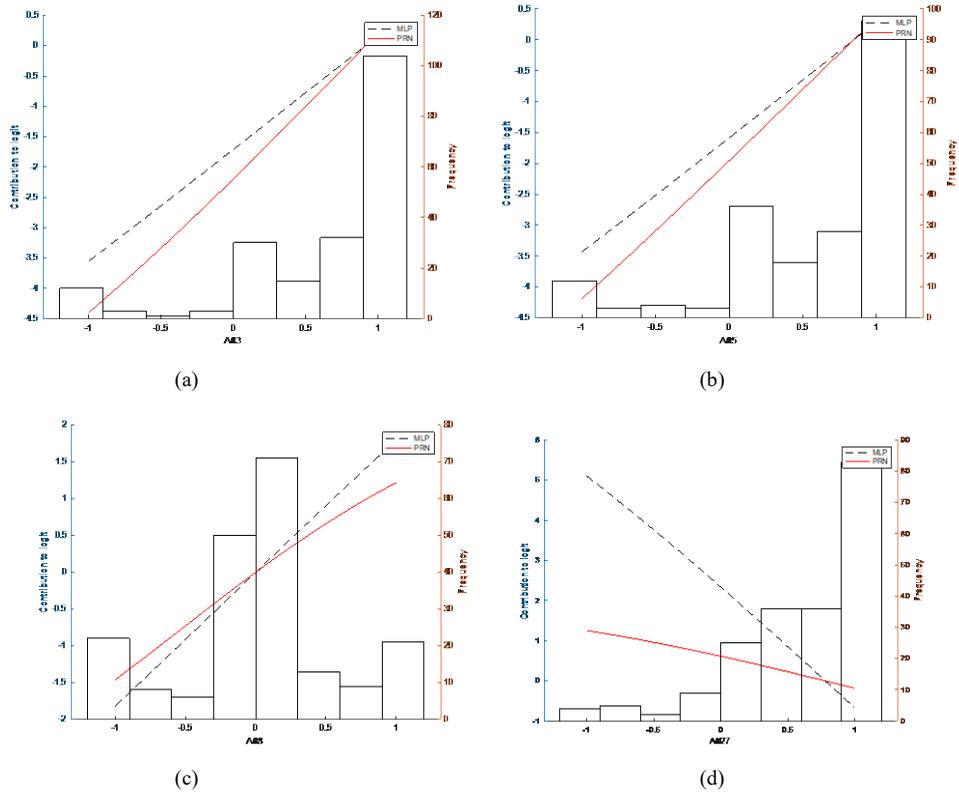

Fig. 4. The Ionosphere model is very stable and includes a discrete variable, Attribute 1, together with the four variables shown.

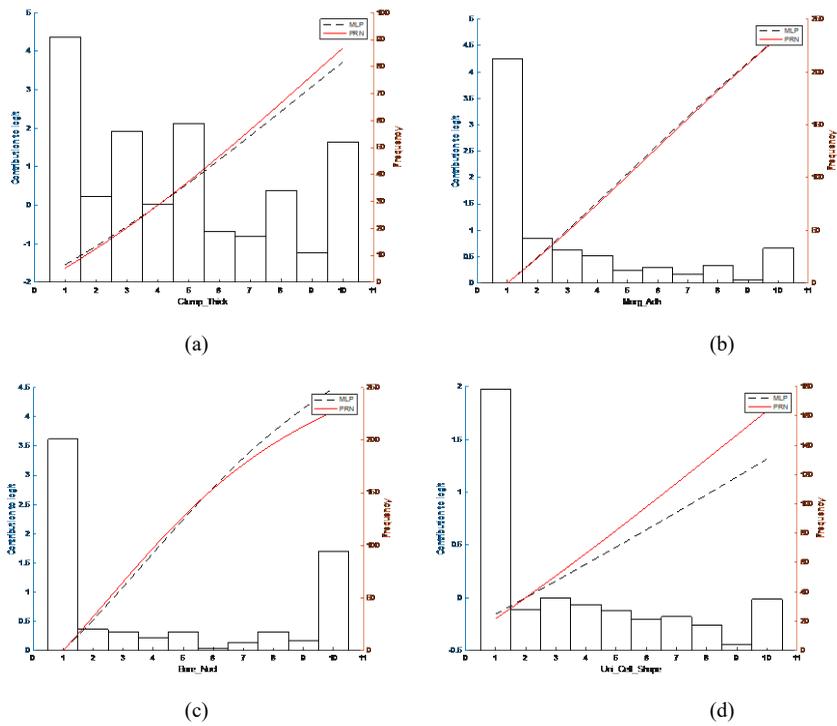

Fig. 5. In the original Wisconsin Breast Cancer data set (WBC-Original) we report four variables that have the highest range of partial response, from six consistently selected.

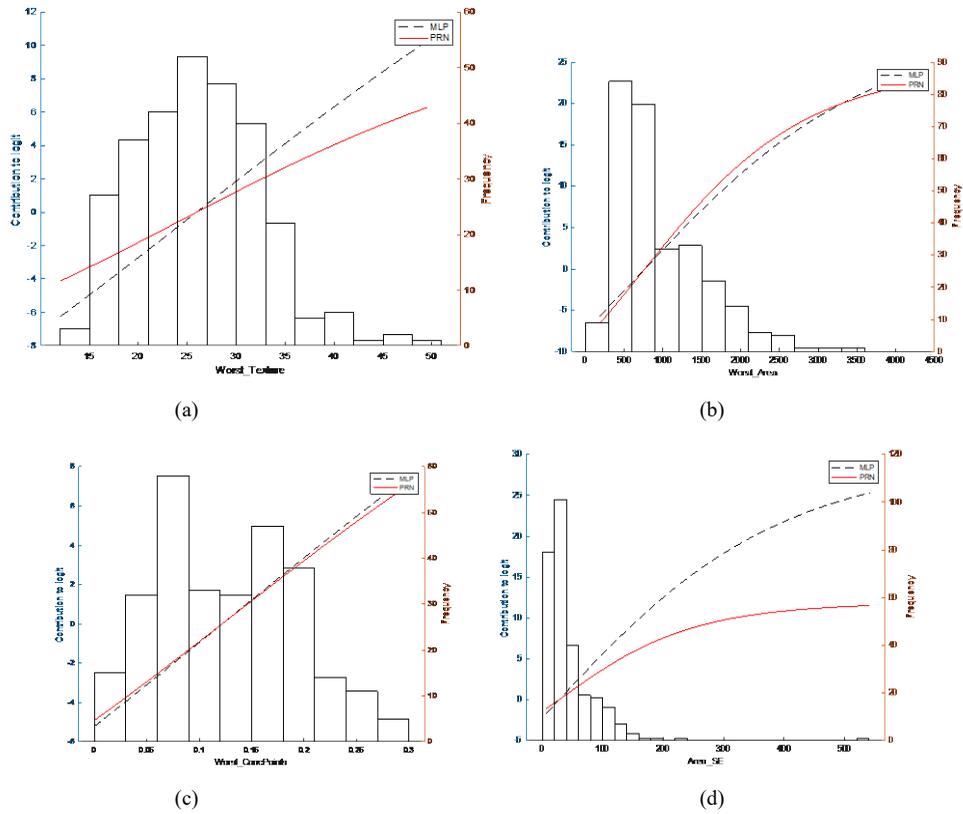

Fig. 6. The second Wisconsin Breast Cancer data set (WBC-Diagnostic) has more non-linear partial functions. Note that in all of the figures the responses are zero at the anchor point, which is the median value.

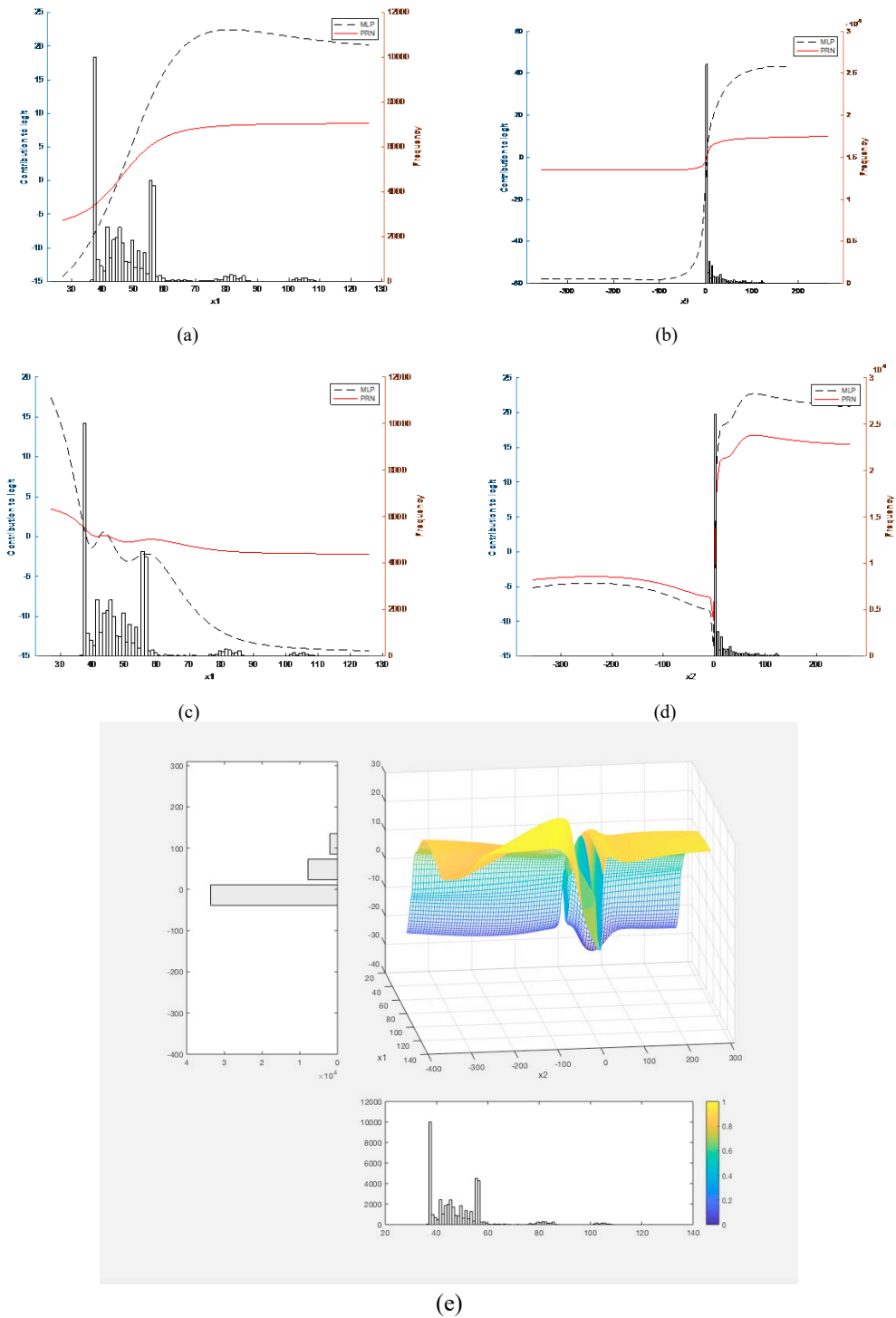

Fig. 7. Partial responses obtained for the Statlog Shuttle dataset using the standard training/test split with sizes n=43,500/14,500 respectively and a prevalence of 21% for class 1 vs. others. (a-b) show two univariates effect for variables $x_1$ and $x_9$ for which the test AUROC is 0.997. (c-e) shows the changes involved in achieving a test AUROC of 1.000, which includes an additional two-way interaction to fit the complexity in the data shown next in fig. 8.

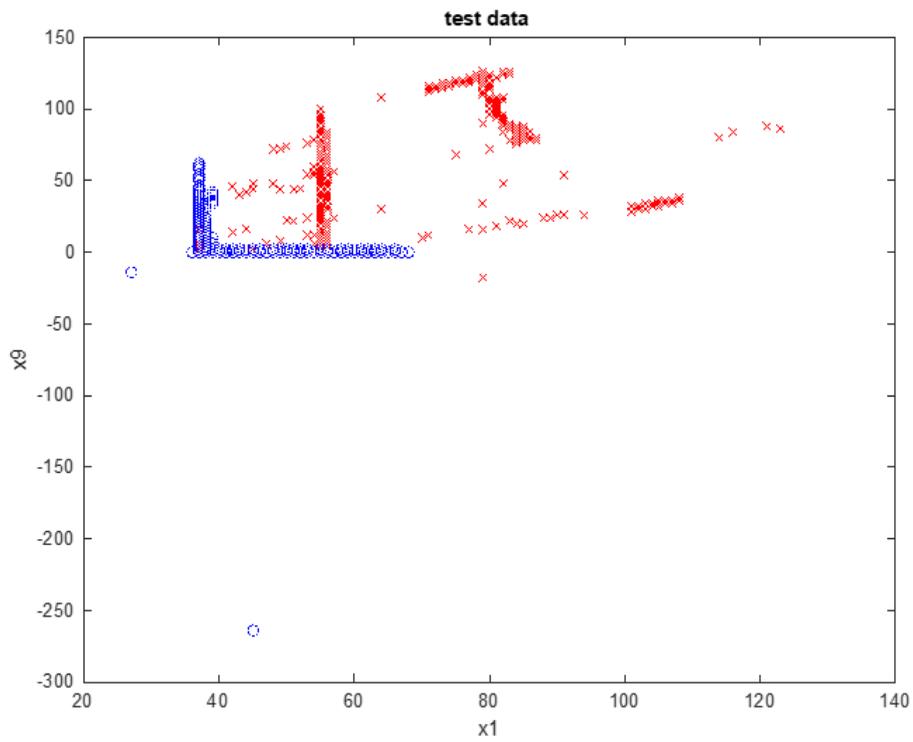

Fig. 8. Plot of the original labelled data for the Statlog Shuttle example,
using only $x_1$ and $x_9$ with class 1 shown in red. This plot explains the results in fig. 7